# A New Random Forest Ensemble of Intuitionistic Fuzzy Decision Trees

Yingtao Ren, Xiaomin Zhu, Kaiyuan Bai, Runtong Zhang, *Senior Member, IEEE*

*Abstract*—Classification is essential to the applications in the field of data mining, artificial intelligence, and fault detection. There exists a strong need in developing accurate, suitable, and efficient classification methods and algorithms with broad applicability. Random forest is a general algorithm that is often used for classification under complex conditions. Although it has been widely adopted, its combination with diverse fuzzy theory is still worth exploring. In this paper, we propose the intuitionistic fuzzy random forest (IFRF), a new random forest ensemble of intuitionistic fuzzy decision trees (IFDT). Such trees in forest use intuitionistic fuzzy information gain to select features and consider hesitation in information transmission. The proposed method enjoys the power of the randomness from bootstrapped sampling and feature selection, the flexibility of fuzzy logic and fuzzy sets, and the robustness of multiple classifier systems. Extensive experiments demonstrate that the IFRF has competitive and superior performance compared to other state-of-the-art fuzzy and ensemble algorithms. IFDT is more suitable for ensemble learning with outstanding classification accuracy. This study is the first to propose a random forest ensemble based on the intuitionistic fuzzy theory.

## I. Introduction

Fuzzy set theory is widely applicable in analyzing uncertainties in data. Crisp classifiers only use the value of the objects, ignoring the relationship between the objects and the concept as a whole. In a fuzzy case, relationships between values and concepts or relationships among values are characterized by the degree of membership. Intuitionistic fuzzy sets can further consider the degree of membership and non-membership by introducing hesitation. It provides a new metric in information processing and makes the decision-making process more similar to the human thinking process. Intuitionistic fuzzy sets can handle uncertainty in different features and have been widely used in machine learning, clustering algorithms, and decision support [1-5].

In classification problems, decision trees can be regarded as one of the most popular classifiers. It has the advantages of fast training speed, small memory consumption, and automatic feature selection. At the same time, it is the most commonly used base classifiers in ensemble learning, such as in bagging, boosting, and random forest [6-9].

However, the crisp decision tree also has certain limitations. For example, insufficient robustness and the relationship between the values and the concepts cannot be revealed intuitively. Many studies have been published regarding this aspect. The most famous one is the fuzzy decision tree (FDT) [10-12]. Fuzzy decision trees bridge the gap between the realm of numbers and the realm of concepts with accurate classification performance [13]. In recent years, there are many articles that have studied the combination of fuzzy theory with decision trees and ensemble algorithms. For instance, Wang et al. [14] proposed a fusing fuzzy monotonic decision tree, which used the fuzzy theory to handle raw data and then input it into a decision tree for classification. Askari et al. [16] used the intuitionistic fuzzy logic in considering cognitive properties of attributes and employed a C4.5 fuzzy decision tree for fraud detection. Segatori et al. [17] developed the fuzzy decision tree for the field of big data computing, which implemented the fuzzy decision tree on the Spark framework for classification. In terms of ensemble learning, Li et al. [15] proposed a fuzzy random forest based on fuzzified features and a boosting method, systematically combining the fuzzy theory and the randomness in the random forest classification process. To further improve the performance of fuzzy ensemble learning, Barsacchi et al. [18] proposed a boosted ensemble algorithm, which was based on week binary fuzzy decision trees and achieved satisfactory classification performance [19]. Apart from fuzzy tree-based classifiers, other types of fuzzy classifiers exist, such as [20] and [21].

In addition, fuzzy random forest has strong capability in handling ambiguity and uncertainty due to the ensembled structure of multiple fuzzy decision trees [22-23]. In recent years, researchers aim to further improve the ability of fuzzy random forests in selecting high-dimensional information and better fitting imperfect data. Bian et al. [24] proposed an enhanced fuzzy random forest method that introduced doubly randomness in the fuzzy random forest, and used a new fuzzy information gain for feature selection. It has been shown to have good processing ability for high-dimensional data. Conn et al. [25] introduced a fuzzy forest algorithm that can provide

Manuscript submitted April 23, 2021; This work was supported in part by the Beijing Natural Science Foundation with grant number L201003 and the National Social Science Foundation of China with grant number 18ZDA086. (*Corresponding author: Xiaomin Zhu.*)

Yingtao. Ren, Xiaomin. Zhu and Kaiyuan. Bai with the School of Mechanical, Electronic and Control Engineering, Beijing Jiaotong University, Beijing 100044, China(e-mail: yingtaoren@bjtu.edu.cn; xmzhu@bjtu.edu.cn; kaiyuanbai@bjtu.edu.cn)

Runtong. Zhang with the School of Economics and Management, Beijing Jiaotong University, Beijing 100044, China(e-mail: rtzhang@bjtu.edu.cn)

less biased feature ranking for correlated high-dimensional data and developed an implementation software. Chen et al. [26] developed a two-layer fuzzy random forest. They adopted the fuzzy C-means [27] to handle high-dimensional emotion features and established the corresponding random forest. Nowadays, methods that combine random forests and fuzzy sets have been applied to immunologic phenotypes [25], fault diagnosis [28], and other classification problems. These advancements have been demonstrated with great potential in theoretical research and applications.

Most of the existing research on fuzzy decision trees and fuzzy random forests is based on the traditional fuzzy sets. Few intuitionistic fuzzy decision tree studies still use fuzzy information as the index of feature selection and ignore the non-membership in the classification process. Moreover, we note that there exist no ensemble algorithms that are designed for intuitionistic fuzzy decision trees. That is, in previous studies, fuzziness was introduced for feature selection purposes, rather than for mimicking human information processing. Also, traditional tree-based methods have only a single degree of membership in transferring information between nodes, resulting in low diversity between classifiers and limited voting strategies. Motivated by the limitations of the existing methods, we propose the intuitionistic fuzzy decision tree (IFDT) by combining intuitionistic fuzzy sets and decision trees, and develop intuitionistic fuzzy random forest (IFRF). In the construction of IFDT, we innovatively introduce hesitation to allow non-membership to be involved in the information transmission of fuzzy decision trees, and propose one type of intuitionistic fuzzy information gain for feature selection in tree nodes. The decision tree inherits the high interpretability of intuitionistic fuzzy sets, and can output to the two dimensions of membership and non-membership. The proposed tree method greatly improves the output selection and randomness. Then, using these characteristics of the IFDT, we design an ensemble learning algorithm, through the random feature selection in the nodes [29] and two different voting schemes to continue increasing the randomness in the decision-making process. In addition, given the classification performance of each random tree on the out of bag dataset (OOB), the voting weight is assigned, which well integrates the high randomness in the decision-making process and the stability of the decision result. Extensive experimental results verify that the proposed IFRF can significantly outperform several state-of-the-art fuzzy classifiers and fuzzy ensemble algorithms.

The rest of this paper is organized as follows. In Section II, we first summarize the fundamental concepts of intuitionistic fuzzy sets and entropy, and then briefly review the existing fuzzy ensembles. In Section III, the proposed approach is described with computational algorithms and illustrative figures. In Section IV, we examine the performance of the proposed decision trees and random forest using multiple benchmark datasets in comparison with alternative methods. We demonstrate the proposed approach has significantly improved accuracy in classification. In Section V, we discuss the advancement and limitations of the proposed method. This method combines the power of randomness in the ensemble and the flexibility of fuzzy logic to handle practical data. Overall, this study is the first to propose a random forest ensemble based on the intuitionistic fuzzy theory.

## II. RELATED WORK

In this section, we summarize the fundamental definitions relating to the intuitionistic fuzzy set and entropy. We also present commonly used intuitionistic fuzzy entropy calculation methods. In addition, classic fuzzy and crisp decision trees and ensemble algorithms are reviewed and briefly discussed.

### A. *Intuitionistic Fuzzy Set and Entropy*

Traditional fuzzy sets are limited to describe the degree to which things belong, called membership. In order to extend the fuzzy set theory, Atanassov [30] first introduced the concept of the intuitionistic fuzzy set, which has been widely adopted in the literature. This concept is theoretically valid with well-defined operations. The intuitionistic fuzzy set not only considers the degree of acceptance, but also the degree of rejection called non-membership, and the degree of uncertainty called hesitation. In this paper, we employ the intuitionistic fuzzy set and the related definitions that are described below.

Let a set X be fixed. One of the most universal generalizations of an intuitionistic fuzzy set A in X has the definition:

$$A = \{< x, u_A(x), v_A(x) > | x \in X\}, \quad (1)$$

where $u_A : X \to [0,1]$ and $v_A : X \to [0,1]$ such that

$$0 \leq u_A(x) + v_A(x) \leq 1. \quad (2)$$

Denote the degree of membership of $x \in A$ as $u_A$ and the degree of non-membership of $x \in A$ as $v_A$. Further, we define the degree of uncertainty as:

$$\pi_A(x) = 1 - u_A(x) - v_A(x). \quad (3)$$

Obviously, for any $x \in X$, we have $0 \leq \pi_A \leq 1$. The value of uncertainty represents the amount of hesitation, which does not belong to membership or non-membership degree.

The fuzzy entropy describes the fuzziness and information content of fuzzy sets. The intuitionistic fuzzy entropy E(x) for each element $x \in A$ is defined as below

$$E_A(x) = \frac{\min\{d_{IFS}(x,M), d_{IFS}(x,N)\}}{\max\{d_{IFS}(x,M), d_{IFS}(x,N)\}}, \quad (4)$$

where M and N are intuitionistic fuzzy elements, $d_{IFS}(x,\cdot)$ is the distances between two intuitionistic fuzzy elements. $M = < u, v, \pi > = < 1, 0, 0 >$ represents fully belonging to set $A$ and $N = < 0, 1, 0 >$ indicates fully not belonging to $A$ [31][32].

Many distance measurements are available, including Euclidean and Hammering distance [33]. Here, we use the normalized Hamming distance. Thus, the $d_{IFS}(x, M)$ and $d_{IFS}(x, N)$ is calculate as:

$$d_{IFS}(x, M) = \frac{(|u_A(x)-1|+|v_A(x)-0|+|\pi_A(x)-0|)}{2}, \quad (5)$$

$$d_{IFS}(x, N) = \frac{(|u_A(x)-0|+|v_A(x)-1|+|\pi_A(x)-0|)}{2}. \quad (6)$$

### B. Fuzzy Decision Trees and Ensemble Algorithms

Decision trees are one of the most commonly used classification methods in machine learning. The generation of decision trees is a recursive process. In every node, we select an appropriate feature to conduct classification. For features with continuous values, crisp decision trees, such as C4.5 and CART, use binary-partition to handle datasets. A dataset is cut into two subsets. The subsets are then divided into different sub-nodes. Similarly, fuzzy decision trees also need to select an feature at each node. However, it can make a sample go down synchronously into multiple sub-nodes with different satisfaction degrees ranging within (0,1].

For the fuzzy decision trees' construction process, the selection of features has many different criteria, including fuzzy information gain [10], fuzzy Gini index [34], minimal ambiguity of the possibility distribution [11], and maximum classification significance of feature contribution [35]. In this paper, we propose a new intuitionistic fuzzy information gain. Details are discussed in Section III.

The ensemble learning methods train multiple learners and combine them to achieve satisfactory performance[36]. In recent literature, many ensemble learning methods have been proposed that use different base classifiers. Commonly used ensemble techniques include bagging, boosting, as well as random forest that has become more popular in recent years. Bagging averages multiple weak classifiers, which are constructed by bootstrapped samples. For boosting, it ensembles weak classifiers by adding only one classifier at a time and weighting each classifier based on its error rate. Random forest method is an extension of bagging, which uses decision trees as the base classifier. In addition to using bootstrapped training data, random forest also randomly selects the set of features at each node.

Compared to the ensemble methods that use crisp decision trees, applying fuzzy decision tree as the base classifier can significantly enhance the robustness and expand the scope of applications. Bonissone[37] proposed a fuzzy random forest, which used results from trees and leaves to vote for the final classification. Barsacchi[18] proposed a boosted ensemble of fuzzy decision trees. Both of the them have shown improvement in accuracy and robustness in numerical studies. Yet, we observe a lack of development in using intuitionistic fuzzy sets for ensemble methods in the current literature. More advanced fuzzy set theory is desired and necessary to obtain better results in classification.

## III. METHOD

In this section, we propose the intuitionistic fuzzy discretizer, intuitionistic fuzzy decision tree, and intuitionistic fuzzy random forest. We also describe the methodological advancements of the proposed method and its implementation in details.

### A. Intuitionistic Fuzzy Discretization

The fuzzy discretizer directly determines the domain of the fuzzy set. We develop an intuitionistic fuzzy discretizer for each continuous feature using K-means method [38] and define strong trapezoid intuitionistic fuzzy partitions.

Common fuzzy sets include triangle, trapezoid, and normal fuzzy sets. Here, we apply trapezoid intuitionistic fuzzy sets. Compared to the traditional trapezoid fuzzy sets, trapezoid intuitionistic fuzzy set has an additional non-membership function (see Fig. 1). We use the hesitation parameter $d_\pi$ to calculate the non-membership function $v_A(x)$:

$$u_A(x) = (1 - u_A(x)) * d_\pi. \quad (7)$$

For example, in Fig. 1, the non-membership function is represented by the red line when $d_\pi = 1$. The non-membership function is represented by the blue line when $d_\pi = 0.5$.

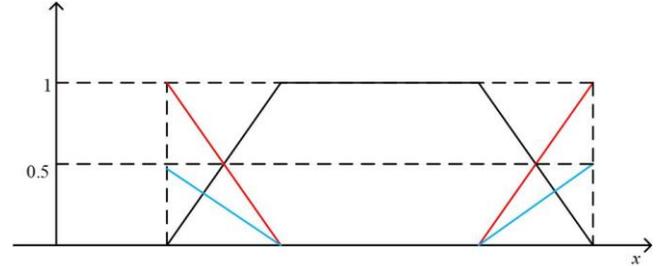

Fig. 1. Trapezoid intuitionistic fuzzy membership and non-membership function (Blue and red lines represent the non-membership function when $d_\pi = 0.5$ and $d_\pi = 1$, respectively)

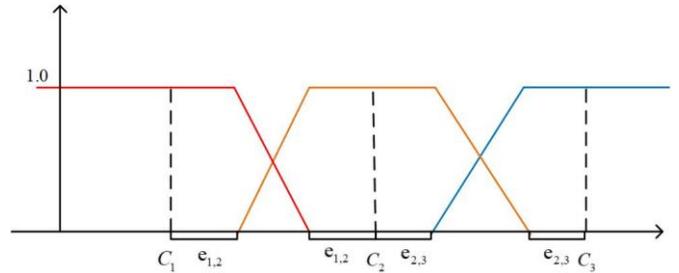

Fig. 2. Trapezoid intuitionistic fuzzy membership function and cluster centers ($e_{I,i+1}$ represents the distance between the i-th trapezoid's center point and the i+1-th trapezoid's adjacent base point)

We defined $C$ as the number of cluster centers in the K-means method. Thus, after clustering certain features in the dataset, we obtain $C$ center points $(c_1, c_2, \dots c_C)$. We define $S$ as the shape parameter, which determines the shape of the membership function. $S$ is calculated as:

$$S = \frac{c_{i+1} - c_i}{e_{i,i+1}} \ (0 < i < C), \quad (8)$$

where $e_{I,i+1}$ $(0 < i < C)$ is denoted as the distance between the i-th trapezoid's center and the i+1-th trapezoid's adjacent base point. Obviously, the trapezoid membership function is approximating to the triangle membership function when S tends to infinity. Here, Fig. 2 vividly demonstrates an example of membership functions when C =3.

The intuitionistic fuzzy discretizer used in this paper is to work on each continuous feature of the dataset and obtain the final intuitionistic fuzzy dataset $D_f$ to train the intuitionistic fuzzy decision tree (IFDT). The process for the discretization is summarized in Algorithm 1.

---

**Algorithm 1:** Pseudocode of the intuitionistic fuzzy discretization process.

For input, the algorithm needs the raw dataset D, the number of clusters C, the shape parameter S, and the hesitation degree $d_\pi$.

The output is the intuitionistic fuzzy dataset $D_f$.

**begin**
    1. Extract the data of one feature from dataset D as the sample set F.
    2. Partition the set F into C clusters using the K-means method and obtain cluster centers.
    3. Intuitionistic fuzzificate the set F towards trapezoid intuitionistic fuzzy set with parameter $d_\pi$ and S.
    4. Repeat step 1, 2, and 3 until all features are processed and the intuitionistic fuzzy dataset $D_f$ is obtained. .
**end**

---

The fuzzy discretizer proposed in this section uses K-means to accurately define clustering centers, and can change the number of membership functions in the domain by changing k-value. It can reveal the underlying information from the data and make full use of the implicit information of the feature values. Only one feature is clustered to obtain the center point at a time, and the class does not participate in the above process. Additionally, it can provide a suitable and diverse intuitionistic fuzzy data set $D_f$ for the subsequent intuitionistic fuzzy decision tree classification.

*B. Construction of Intuitionistic Fuzzy Decision Trees*

Here, we propose an intuitionistic fuzzy information entropy. Different from the information entropy used by C4.5 decision tree, we calculate the intuitionistic fuzzy entropy of each fuzzy partition of each sample according to (4).

Assume that the proportion of the feature F's intuitionistic fuzzy entropy of samples that belongs to the k-th class in the current intuitionistic fuzzy sample set $D_f$ is $P_k^F$, where k = 1,2…, |y|. The $P_k^F$ is calculated as:

$$P_k^F = \frac{\chi(s)*E_F(s),\quad s\in \text{the k th class of } D_f}{\chi(s)*E_F(s),\quad s\in D_f} \qquad (9)$$

Assume that feature F has C fuzzy partitions $\{a^1, a^2, …, a^c\}$. If F is used to split node, C child nodes are generated. The c-th child node contains all samples in $D_f^{parent}$, of which the membership degree is greater than zero on the fuzzy partition $a^c$. The set is denoted as $D_f^c$. We can compute intuitionistic fuzzy information entropy of the set $D_f^c$ using (10).

$$\text{IEnt}(D_f^{parent}, F) = -\sum_{k=1}^{|y|} P_k^F \log_2 P_k^F \qquad (10)$$

Consider that the sum values of intuitionistic fuzzy entropy in different child nodes are different. Weights are assigned to each child node, which are the sum of intuitionistic fuzzy entropy in $D_f^c$ divided by the sum of intuitionistic fuzzy entropy in $D_f^{parent}$. The larger the weight, the greater the influence of the child node. Consequently, the "intuitionistic fuzzy information gain" obtained by splitting the intuitionistic fuzzy sample set $D_f^{parent}$ with feature F can be calculated as:

$$\text{IGain}(D_f^{parent}, F) = \text{IEnt}(D_f^{parent}, F) - \sum_{c=1}^C \frac{|D_f^c|}{D_f^{parent}} \text{IEnt}(D_f^c, F) \qquad (11)$$

We note that the mean value of the intuitionistic fuzzy entropy of all features is used when calculating the intuitionistic fuzzy information entropy $\text{IEnt}(D_f^{parent})$ of the sample set $D_f^{parent}$ in the splitting parent node.

At the beginning of the IFDT construction, a root node is created with all samples in $D_f$. We denote $\chi_{t,N}(s)$ as the degree of membership to indicate the condition that leads to the node N of the tree t, $s \in D_f$. The calculation of $\chi_{t,N}(s)$ is described as follows:

- At the root node, each sample s has been allocated with an initial value 1, as $\chi_{t,root}(s) = 1$.
- In the process of node splitting, samples may belong to one or more child nodes due to the overlapping of the intuitionistic fuzzy sets. Samples are descended to a child node associated with the membership degree greater than 0 as $u_{fuzzy\_partition}(s) > 0$. Thus, the membership value $\chi_{t,childnode}(s)$ is calculated as:

$$\chi_{t,childnode}(s) = \chi_{t,node}(s) \times u_{fuzzy\_partition}(s). \qquad (12)$$

Similarly, the non-membership value $\psi_{t,childnode}(s)$ is calculated as:

$$\psi_{t,childnode}(s) = \psi_{t,node}(s) \times v_{fuzzy\_partition}(s). \qquad (13)$$

- If the sample s has a missing value on a feature that is used to split node. The value $\chi_{t,childnode}(s)$ is calculated as:

$$\chi_{t,childnode}(s) = \frac{\chi_{t,node}(s)}{C} \qquad (14)$$

As the same as the traditional decision trees, the proposed IFDT is constructed by a recursive node-splitting process,

which searches for the split that maximizes the intuitionistic fuzzy information gain until meeting a series of stopping conditions. This learning process is summarized in Algorithm 2.

---

**Algorithm 2:** Algorithm for constructing the proposed IFDT.

For input, we have the algorithm intuitionistic fuzzy dataset $D_f$, the stopping parameters $d_\beta$ and $n_\alpha$, The output is an intuitionistic fuzzy decision tree IFDT.

---

**begin**
1. Create a root and start with the samples in $D_f$, set $\chi_{root}(s) = 1$.
2. Let M be the number of features in $D_f$. Let C be the number of fuzzy partitions.
3. Select a feature to split the current node N.
   3.1 Calculate the intuitionistic fuzzy information gain for each feature using the intuitionistic fuzzy entropy E(s) of each samples in node N.
   3.2 Select the feature with the maximum intuitionistic fuzzy information gain to split.
4. Generate C child nodes for the node N in accordance with the number of fuzzy partitions. If $u_{fuzzy\_partition}(s) > 0$, the sample is assigned to the corresponding child node. One sample may be assigned to multiple child nodes.
5. Repeat step 3 and 4 for each node until the stopping condition is reached.

**end**

---

In this paper, the stopping criteria in Algorithm 2 are:
- The tree's depth reaches the maximum depth $d_\beta$.
- Current node contains samples less than a threshold $n_\alpha$.
- Current node contains samples belong to the same class.
- The set of alternate features is empty.

When any one of the above conditions is satisfied, the split of the node stops and the node is marked as a leaf node with a certain class that have the largest proportion of $\chi_{t,leaf} - \psi_{t,leaf}$. Otherwise, C child nodes are generated, selecting the split that maximizes the intuitionistic fuzzy information gain.

Comparing the proposed intuitionistic fuzzy decision tree to the traditional fuzzy decision tree, two major advancements are achieved. First, we propose a new intuitionistic fuzzy information gain as the splitting criterion and extend the feature selection method using the intuitionistic fuzzy entropy. Second, we adopt the non-membership to the information transmission of the decision trees and intuitively associate value with concept. These advancements upgrade the decision tree technique to more effectively mimic human thinking process. The proposed IFDT also has strong generalization ability. Moreover, due to the addition of one dimension of output, IFDT is more suitable for being used as the base classifier for ensemble learning. These advanced characteristics are verified in the section IV.

*C. Intuitionistic Fuzzy Random Forest Learning and Classification*

In order to better use the two-dimensional output of IFDT, we propose the intuitionistic fuzzy random forest (IFRF) by strategically combining the IFDT, and design two different voting schemes for decision making.

In the construction of IFRF, similar to random forest algorithm, we used the intuitionistic fuzzy random decision tree as the base classifier, which adds random selection of features in each node. During the construction of intuitionistic fuzzy random decision tree, suppose there are m candidate features in a node. Different from IFDT, the proposed decision tree randomly selects a subset of $\log_2 m$ features from the features set of the node, and then select the optimal feature from this subset to generate child nodes. The rest of the construction is exactly the same as IFDT. The detailed process of constructing IFRF is summarized in Algorithm 3.

---

**Algorithm 3:** Algorithm for constructing the proposed IFRF.

For input, we have the intuitionistic fuzzy dataset $D_f$, number of trees $n_\gamma$, the stopping parameters $d_\beta$ and $n_\alpha$, The output is an intuitionistic fuzzy random forest IFRF.

---

**begin**
1. Extract a random sample set |D| with replacement from the dataset $D_f$ and obtain the out of bag samples set OOB.
2. Input |D|, stopping parameters and construct an intuitionistic fuzzy random decision tree.
3. Use OOB dataset to test the base random decision tree and obtain the classification error rate.
4. Repeat steps 1, 2, and 3 until the number of trees reach $n_\gamma$.
5. Assign the weight $w_t$ ($0 < t \leq n_\gamma$) to each base tree according to its classification error rate on the OOB dataset.

**end**

---

|        | tree 1            | tree 2            | tree 3            |
|--------|-------------------|-------------------|-------------------|
| leaf 1 | class1 , $L_{1,1}$ | class1 , $L_{2,1}$ | class1 , $L_{3,1}$ |
| leaf 2 | class2 , $L_{1,2}$ | null              | class1 , $L_{3,1}$ |
| leaf 3 |                   | class2 , $L_{2,1}$ |                   |
| leaf 4 |                   | class1 , $L_{2,1}$ |                   |
| weight | $w_1$             | $w_2$             | $w_3$             |

Fig. 3. A result matrix example of a IFRF that has three trees

In the first step of Algorithm 3, the resampling with replacement is consistent with the traditional random forest method using bootstrapping[39]. As a result, approximately 63.2% of the samples are extracted to form the dataset |D|. The rest of approximately 36.8% samples that were not included are called the out of bag (OOB) dataset. The OOB dataset are

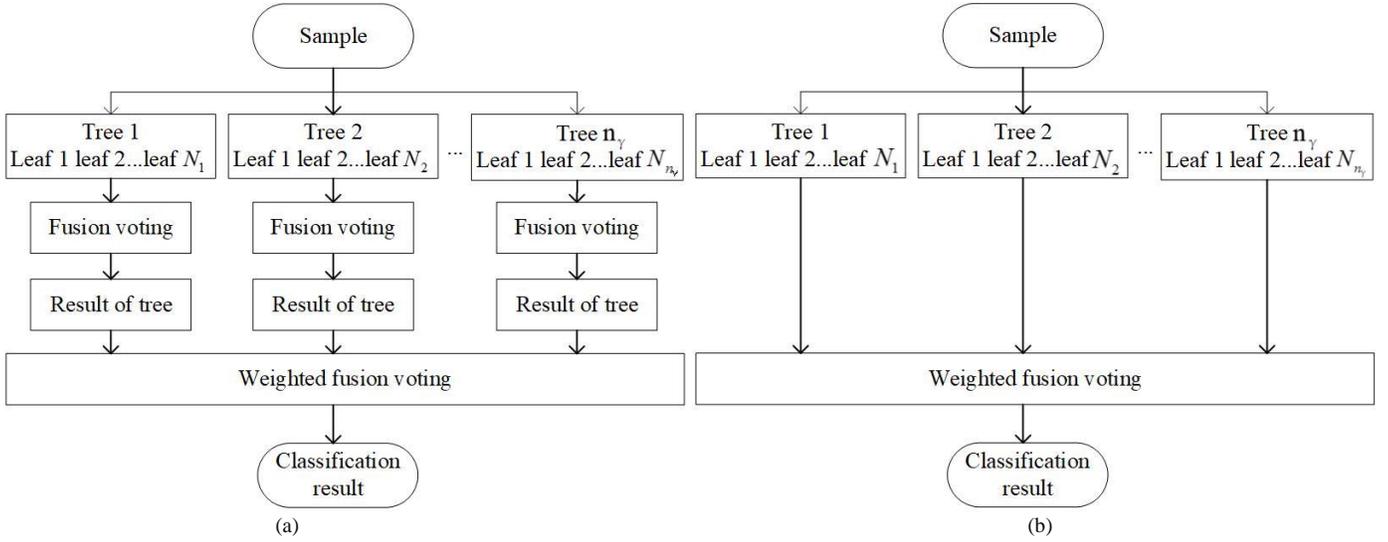

Fig. 4. Two voting schemes for the IFRF: (a) tree-based voting scheme 1. (b) leaf-based voting scheme 2

used to test the intuitionistic fuzzy random decision tree and assign weights to the random decision trees based on the testing results in the step 5. In the step 5 of Algorithm 3, the weights of t-th tree in IFRF is computed as:

$$w_t(err_t) = \begin{cases} 1, & \text{if } err_t \leq (emin + marg) \\ \frac{(emax + marg) - err_t}{(emax - emin)}, & \text{if } (emin + marg) \leq err_t \leq (emax) \end{cases} \quad (15)$$

where the $err_t$ is the error rate of the t-th tree in IFRF, which obtained as $\frac{errors_{(OOB_t)}}{size_{(OOB_t)}}$. The $errors_{(OOB_t)}$ is the number of classification errors of t-th tree that uses $OOB_t$ dataset as test set, and the $size_{(OOB_t)}$ is the number of samples of the $OOB_t$ [37]. The emax is the maximum error rate in the trees of IFRF, which is calculated as:

$$emax = \max_{t=1,\dots n_\gamma} \left\{ \frac{errors_{\{OOB_t\}}}{size_{\{OOB_t\}}} \right\} \quad (16)$$

Similarly, the emin is the minimum error rate in the trees of IFRF and $marg = (emax - emin)/4$.

When classifying a new sample, we assume that the t-th tree generates $N_t$ leaves. If a sample reaches a certain leaf in the classification process, the leaf is assigned to a voting weight $L_{t,leaf} = \chi_{t,leaf} - \psi_{t,leaf}$. The leaves that have not been reached are null and do not participate in voting. A result matrix example is shown in Fig. 3

The classification voting for the final decision follows the majority voting principle. We develop two voting strategies: tree-based voting (scheme 1) and leaf-based voting (scheme 2). The scheme 1 is to perform a fusion voting for the leaf voting weights within each tree to obtain a classification result of each tree, and then conduct a weighted fusion voting for these results to generate the final decision of the IFRF. The scheme 2 is to directly conduct a weighted fusion voting by combining the information generated by all leaves. We plot the two-voting scheme in Fig. 4.

The fusion voting in scheme 1 is to add up the leaves' voting weight of each class on each tree, obtain the class with the largest weight, and mark the tree result as the class. Then, the voting weights of the trees of each class are added to obtain the class with the largest sum, and the class is outputed as the final ensemble classification result. The weighted fusion voting in scheme 2 is to multiply the voting weight of each leaf by the voting weight of its belonging tree and add them according to the class of the leaves. We obtain the class with the largest weight and output this class as the final ensemble classification result.

Through proper voting schemes, IFRF integrates the advantages of intuitionistic fuzzy set theory into the random forest's classification process, making the final fused voting results more accurate. In addition, by considering the role of membership and non-membership in information transmission, the randomness in the IFRF classification process is greatly increased. In the process of integrating various base learners, OOB is used to weight them, which increases the stability of voting. The above strategies combine the advanced characteristics of intuitionistic fuzzy theory, include the important role of $\chi_{t,leaf}$ in voting, and enhance the performance of the proposed IFRF in classification accuracy.

## IV. EXPERIMENT

The evaluation of classification algorithms is a delicate job that requires comparisons using multiple datasets. In order to examine the performance of our proposed classification algorithms and ensure the reproducibility of the results, we select 23 commonly used classification datasets from Uci and Keel and adopt cross validation for evaluation [40]. These datasets are extensively used in the literature for many ensembles learning models. Furthermore, the used datasets cover various aspects of general classification problems, involving different numbers of samples, features, balance levels, and real/integer/numerical features in different ratios. The ultimate goal we aim to achieve is to comprehensively assess the classification performance of the IFDT classifier and the IFRF ensemble algorithm, in comparison with other fuzzy classification methods. The selected datasets are listed in Table I with the numbers of features (real(R), integer(I), categorical(C)), the numbers of classification class, and the imbalance ratio (IR). The IR is the ratio of the class with the largest sample size to the class with the smallest sample size.

To evaluate of a classifier, accuracy is an appropriate and intuitive measurement to quantify performance levels. Thus, we design two series of experiments to investigate the accuracy of a single IFDT and IFRF ensemble algorithm respectively. In the first series of experiments, we focus on the comparison of IFDT with other popular base classifiers, including C4.5, CART, and MLP[41]. The purpose is to properly analyze the performance of our proposed base classifier IFDT. Similarly, in the second series of experiments, we aim to compare the IFRF ensemble method with other popular ensembles in the literature.

TABLE I
DATASETS AND THEIR CHARACTERISTICS

| Dataset | Samples | Feature (R,I,C) | Class |
|---|---|---|---|
| appendicitis (APP) | 106 | 7 (7,0,0) | 2 |
| australian (AUS) | 690 | 14 (3,5,6) | 2 |
| bands (BAN) | 365 | 19 (13,6,0) | 2 |
| dermatology (DER) | 358 | 34 (0,34,0) | 6 |
| glass (GLA) | 214 | 9 (9,0,0) | 7 |
| hayes (HAY) | 160 | 1 (0,4,0) | 3 |
| iris (IRI) | 150 | 4 (4,0,0) | 3 |
| magic (MAG) | 19020 | 10 (10,0,0) | 2 |
| mammographic (MAM) | 830 | 5(0,5,0) | 2 |
| newthyroid(NEW) | 215 | 5 (4,1,0) | 3 |
| ring (RIN) | 7400 | 20 (20,0,0) | 2 |
| segment (SEG) | 2310 | 19 (19,0,0) | 7 |
| tae (TAE) | 151 | 5 (0,5.0) | 3 |
| vehicle (VEH) | 846 | 18 (0,18,0) | 4 |
| vowel (VOW) | 990 | 13 (13,0,0) | 11 |
| wdbc (WDC) | 569 | 30 (30,0,0) | 2 |
| wine (WIN) | 178 | 13 (13,0,0) | 3 |
| wisconsin(WIS) | 683 | 9(0,9,0) | 2 |
| german credit(GER) | 1000 | 20(0,7,13) | 2 |
| thyroid (THY) | 3772 | 21(15,6,0) | 3 |
| breast-cancer(BC) | 286 | 9(0,1,8) | 2 |
| ionosphere (ION) | 351 | 33(32,1,0) | 2 |
| pima (PIM) | 768 | 8(8,0,0) | 2 |

In the following, we test the proposed methods separately. The two series of experiments follow the standard procedure. First, fit a dataset-specific tuning parameter by each method in each dataset. Here, we use a nested 5-fold cross-validation testing sets to select hyperparameters. Subsequently, we test IFDT and IFRF respectively. In order to fairly compare with other classification methods in the literature, we follow Bonissone [37] and Barsacchi [18], running cross-validation [42] test for IFDT and IFRF. Our reference implementation of IFDT and IFRF are developed in Python 3.7, and it is used in all experiments described hereafter.

In section IV.A, we evaluate the accuracy of our base fuzzy classifier IFDT and other fuzzy and crisp base classifiers. In Section IV.B, the proposed IFRF is analyzed and compared with others newest fuzzy ensemble classification methods.

### A. Evaluation of Intuitionistic Fuzzy Decision Trees

In this section, we compare the proposed IFDT with other general decision tree-based classifiers and fuzzy classifiers.

We consider two commonly used crisp decision tree C4.5 [44]and CART [45] that use information gain ratio and Gini index as the criteria of feature selection in nodes respectively. Fuzzy decision trees incorporate fuzzy theory into the traditional crisp information gain with the fuzzy information gain, which makes fuzzy decision trees widely applicable in other fields. Multilayer Perceptron (MLP) [47] and radial base function (RBF) [48] are two general network classifiers. Bayes classifier is a popular classifier, which is based Bayesian decision theory. The above six classifiers are often used as the base classifiers for many ensemble learning methods. For example, Ada-boosting and Bagging. We also consider grey relational analysis GRA [46] and CIGRA classifier [41] which adds to fuzzy theory for more comprehensive comparisons.

The selection of the hyperparameters in the classifier is mainly based on the grid search method and 5-fold inner cross-validation. For IFDT, $C \in [2:7]$ with step 1, $S \in [2,3,4,5,7,9,10000]$, $d_\pi \in [0.5:0.9]$ with step 0.1, $n_\alpha \in [5:20]$ with step 5, $d_\beta \in [2:20]$ with step 2.

In order to fairly compare with each classifier, we use the reported results from [41]. The accuracy results are obtained by a 10-fold cross validation testing strategy. The percentage of the average classification accuracy for all classifies are reported in Table II.

We observe in Table II that CIGRA, IFDT and RBF perform better in accuracy. IFDT and RBF both perform the best on three datasets, and CIGAR performs the best on two datasets. In addition, we calculate the mean rank of error rate of each classifier on eight datasets. We observe that in terms of the mean rank, the CIGRA classifier performers the best with a mean rank of 2.80, followed by the IFDT classifier with a mean rank of 3.65, the two neural network classifies RBF and MLP.

In order to further explore the difference between the results of each classifier, we perform a non-parametric test on the error rate results. We choose Friedman test [49] that is widely used in comparing algorithm performance, and the Friedman test results are reported in Table III (Friedman test results are obtained using SPSS 22). In this paper, we choose the significance p-value as 0.05. From Table III, the Friedman test produce a p-value = 0.019, lower than 0.05, indicating the null hypothesis of statistical equivalence is rejected. Therefore, we conclude that there is a significant difference between the accuracy of classifiers.

TABLE II
ACCURACY RESULT OF THE TEST OF IFDT AND OTHER CLASSIFIES

| Datasets | Classifiers | | | | | | | | |
|---|---|---|---|---|---|---|---|---|---|
| | IFDT | GRA | CIGRA | MLP | C4.5 | RBF | Bayes | Cart | FDT |
| APP | 88.02 | 86.00 | **88.70** | 85.80 | 84.9 | 80.20 | 83.00 | 84.90 | 88.10 |
| BC | **78.43** | 72.80 | 74.30 | 73.50 | 73.9 | 72.40 | 69.30 | 71.40 | 77.75 |
| WIS | **96.88** | 96.20 | 96.80 | 96.50 | 94.7 | 96.60 | 96.40 | 94.40 | 96.76 |
| GER | 71.03 | 73.00 | 74.20 | 71.60 | 73.5 | **75.70** | 70.40 | 73.90 | 68.39 |
| GLA | **74.45** | 57.40 | 63.20 | 68.70 | 65.8 | 47.70 | 71.80 | 63.60 | 71.36 |
| ION | 90.71 | 88.50 | 92.60 | 92.00 | 90.9 | **94.60** | 85.50 | 89.50 | 88.66 |
| IRI | 96.00 | 95.70 | 96.10 | 96.00 | 94.00 | **98.00** | 94.70 | 92.00 | 95.33 |
| PIM | 75.61 | 74.90 | **76.20** | 75.80 | 72.7 | 75.70 | 72.20 | 74.70 | 74.26 |
| WIN | 97.65 | 93.30 | 96.20 | **98.30** | 93.3 | 94.90 | 94.40 | 87.60 | 97.14 |
| THY | 93.74 | 97.50 | 98.20 | 94.30 | 99.7 | 95.50 | **99.70** | 99.60 | 95.28 |

TABLE III
RESULTS OF THE FRIEDMAN TEST ON THE ACCURACY OBTAINED BY IFDT, GRA, CIGRA, MLP, C4.5, RBF, BAYES, CART AND FDT

| Classifier | Mean Rank | p-value |
|---|---|---|
| **IFDT** | 3.65 | **0.019** |
| GRA | 6.05 | |
| CIGRA | 2.80 | |
| MLP | 4.25 | |
| C4.5 | 5.65 | |
| RBF | 4.60 | |
| Bayes | 6.55 | |
| Cart | 6.55 | |
| FDT | 4.90 | |

TABLE IV
RESULTS OF THE PAIRWISE FRIEDMAN TESTS
(SIGNIFICANCE LEVEL = 0.05)

| Classifier | Mean Rank | p-value | Hypothesis |
|---|---|---|---|
| CIGRA | 2.80 | 0.527 | Not reject |
| IFDT | 3.65 | - | - |
| MLP | 4.25 | 0.739 | Not reject |
| RBF | 4.60 | 1.000 | Not reject |
| FDT | 4.90 | 0.058 | Not reject |
| C4.5 | 5.65 | 0.206 | Not reject |
| GRA | 6.05 | 0.058 | Not reject |
| Bayes | 6.55 | 0.011 | Rejected |
| Cart | 6.55 | 0.058 | Not reject |

In order to further compare the performance of each classifier with IFDT, we choose the proposed IFDT as the control classifier, and conduct pairwise Friedman tests with other classifiers. The results are shown in Table IV. From Table IV, we conclude that there is no significant difference in performance between IFDT and CIGRA, MLP, RBF, FDT, C4.5, GRA and CART. The proposed IFDT is significantly better than the Bayes classifier. However, using the mean rank as the evaluation measurement, IFDT ranks as the second among nine classifiers, which demonstrates that the proposed IFDT has competitive performance and is an effective classifier.

From the above experiments, we note that the intuitionistic fuzzy decision tree classifier proposed in this paper is an effective classifier with competitive performance in accuracy among other eight classifiers. Furthermore, since it considers the degree of hesitation and is able to output decision information in two dimensions, the proposed IFDT is practically applicable to many other fields to produce accurate classification results and more suitable for ensemble learning.

*B. Performance Analysis for the IFRF Ensemble*

In this section, we compare the performance of the proposed IFRF with four state-of-the-art fuzzy classifiers and ensemble algorithms and classic random forest (RF). FDT-Boost [18] is an ensemble algorithm, which uses fuzzy binary decision trees as a multi-class base classifies. This type of trees is constrained their depth to maintain compact while not reducing the classification accuracy. FURIA [43] is an extended algorithm based on the RIPPER algorithm, which classifies by learning conventional rules and unordered rule sets. Moreover, we also choose two latest fuzzy decision trees: FBDT and FMDT [17]. Both of them adopt strong fuzzy partitioning in the domains of numeric features. We note that FBDT is fuzzy binary decision tree and FMDT is fuzzy multi-way decision tree.

We employ a 5-fold cross-validation method in the performance evaluation of the IFRF and other four algorithms. The hyperparameter selection method in this experiment is the same as Section IV.A. For IFRF, $C \in [2:7]$ with step 1, $S \in [2,3,4,5,7,9,10000]$, $d_\pi \in [0.5:0.9]$ with step 0.1, $n_\alpha \in [5:20]$ with step 5, $d_\beta \in [2:20]$ with step 2 and $n_\gamma = 600$.

In practice, it is not necessary to tune all the hyperparameters to perform a grid search. We conduct a grid search only on selected hyperparameters that have a large impact on the model, and then simply adjust the remaining hyperparameters or set them as default values. The influence of specific hyperparameters on the model and the default values are discussed in the appendix. For the other four algorithms, they also select dataset-specific hyperparameters by tunning.

To make the experiment result more comparable, the result reported in [18] is used and the complete experimental results are reported in Table V.

At the first glance, the best classifier is the proposed IFRF, because it has the highest accuracy on two thirds of the test datasets. Its performance on the remaining datasets is also close to the best performance. In order to quantify this observation, we also conduct a non-parameter statistical test based on the error rates of each classifier on the test datasets. The specific testing process is the same as the section IV.B. The results of the total Friedman test are reported in Table VI, and the results of the paired-Friedman tests are reported in Table VII. It is worth noting that, to make the conclusion more rigorous, we use the Holm-method [50] to adjust the significance level for multiple comparison.

TABLE V
MEAN ACCURACY RESULTS OBTAINED BY IFRF, FDT-BOOST, FURIA, FBDT AND FMDT

| Datasets | Algorithms | | | | | | | | | |
|---|---|---|---|---|---|---|---|---|---|---|
| | IFRF | | FDT-Boost | | FURIA | | FBDT | | FMDT | |
| | mean | std | mean | std | mean | std | mean | std | mean | std |
| APP | **90.56** | 5.21 | 85.84 | 7.38 | 90.38 | 1.10 | 87.75 | 2.24 | 84.07 | 6.68 |
| AUS | 86.67 | 2.22 | **86.87** | 1.87 | 82.44 | 6.99 | 84.49 | 1.98 | 85.36 | 3.41 |
| BAN | 73.48 | 4.51 | **75.31** | 3.75 | 73.26 | 4.66 | 65.17 | 3.10 | 68.55 | 3.09 |
| DER | **98.09** | 1.86 | 98.06 | 2.40 | 81.7 | 1.46 | 81.70 | 1.49 | 93.3 | 3.21 |
| GLA | 78.47 | 7.53 | 75.70 | 6.98 | **82.62** | 3.94 | 66.79 | 5.22 | 72.39 | 5.42 |
| HAY | **84.37** | 3.43 | 83.75 | 4.59 | 75.00 | 2.53 | 73.75 | 4.24 | 63.75 | 7.02 |
| IRI | 97.33 | 2.49 | 94.67 | 2.67 | **98.00** | 0.42 | 96.00 | 3.89 | 94.00 | 4.42 |
| MAG | **86.03** | 0.25 | 85.18 | 0.65 | 85.49 | 0.12 | 85.63 | 0.73 | 80.03 | 0.77 |
| MAM | **84.07** | 2.52 | 83.96 | 1.58 | 79.04 | 6.32 | 83.23 | 2.39 | 83.93 | 2.46 |
| NEW | **98.14** | 2.71 | 96.74 | 3.15 | 97.67 | 0.59 | 93.02 | 2.94 | 93.49 | 1.70 |
| RIN | **95.89** | 0.99 | 95.54 | 0.32 | 83.7 | 6.91 | 92.55 | 0.65 | 91.28 | 1.08 |
| SEG | **97.48** | 0.70 | 97.4 | 0.64 | 84.52 | 4.80 | 96.36 | 0.57 | 95.84 | 0.86 |
| TAE | **57.59** | 4.09 | 52.32 | 4.90 | 48.74 | 3.47 | 49.01 | 3.26 | 51.61 | 4.87 |
| VEH | **74.23** | 2.15 | 74.11 | 1.89 | 71.09 | 2.05 | 71.40 | 1.89 | 70.33 | 1.71 |
| VOW | **97.98** | 0.90 | 95.55 | 1.25 | 79.33 | 1.86 | 79.6 | 1.94 | 94.44 | 1.59 |
| WDC | 97.54 | 0.65 | 97.01 | 1.43 | **98.42** | 0.60 | 94.38 | 1.62 | 94.55 | 1.18 |
| WIN | **99.44** | 1.11 | 98.87 | 1.38 | 98.43 | 0.97 | 91.51 | 5.73 | 94.38 | 2.52 |
| WIS | 97.42 | 1.07 | **97.66** | 0.84 | 95.20 | 7.34 | 95.32 | 1.06 | 95.03 | 1.23 |

TABLE VI
RESULTS OF THE FRIEDMAN TEST ON THE ACCURACY OBTAINED BY IFRF, FDT-BOOST, FURIA, FBDT AND FMDT

| Algorithms | Mean Rank | p-value |
|---|---|---|
| **IFRF** | 1.33 | **<0.001** |
| FDT-Boost | 2.33 | |
| FURIA | 3.42 | |
| FBDT | 3.86 | |
| FMDT | 4.06 | |

TABLE VII
RESULTS OF THE PAIRWISE FRIEDMAN TEST ON IFRE AND OTHER FOUR ALGORITHMS WITH HOLM METHOD (SIGNIFICANCE LEVEL = 0.05)

| Algorithms | Holm | p-value | Hypothesis |
|---|---|---|---|
| **FBDT** | 0.0125 | **<0.001** | Rejected |
| **FMDT** | 0.0167 | **<0.001** | Rejected |
| **FDT-Boost** | 0.25 | **0.005** | Rejected |
| **FURIA** | 0.05 | **0.005** | Rejected |

According to Table VI, the test result of the IFRF algorithm that ranks the first in accuracy has a p-value <0.001 and the null hypothesis of statistical equivalence is rejected. Therefore, we conclude that there is a significant difference between each classifies, and the IFRF preforms the best in classification accuracy. The subsequent pairwise Friedman test with the Holm adjustment also proves this conclusion.

The above two series of experiments comprehensively demonstrate the feasibility of integrating intuitionistic fuzzy theory into decision tree algorithm. The IFDT classifier has competitive performance with other general classifiers, and the IFRF algorithm with IFDT as the base classifier has obviously superior performance in the comparison with other state-of-the-art algorithms.

We demonstrate the superior performance in classification accuracy and great integration potentials using various experiments. The additional dimension information in outputs increases its ensemble performance, making the IFRF outperform in accuracy. These results confirms that the combination of the intuitionistic fuzzy sets and decision tree ensemble algorithms is effective and efficient.

The time complexity of IFRF is lower than that of traditional random forests. Since IFRF supports multi-split, when multi-split is conducted, the space complexity will be larger than that of traditional random forests, while the storage space size remains acceptable under a regular laptop configuration. For a detailed analysis, see Appendix B.

*C. Performance Analysis for the IFRF in Noise Dataset in comparison with crisp ensemble learning algorithms*

A significant advantage of the ensemble learning algorithm is its robustness; fuzzy models are considered to also have this property. In order to further study the robustness of the combination of intuitionistic fuzzy sets and random forest, we select two ensemble learning algorithms that use crisp decision trees as their base learner, random forest (RF) and AdaBoost as comparisons. The robustness of these algorithms is evaluated using two types of label noise: moderate noise (10%) and heavy noise (30%), with results from original data (no noise) serving as a benchmark. Before the experiment, we randomly shuffle the labels of a certain number of samples in the training dataset according to the noise level to obtain noise datasets. Both random forest and AdaBoost methods use the algorithms in scikit-learn, the most popular machine learning Python library, among which AdaBoost uses SAMMA. Both algorithms use grid search to tune the optimal model in each dataset. The number of trees included in each algorithm is set to be 100. Other hyperparameters are tuned in the same process as the IFRF in section IV.B. We also employ a 5-fold cross-validation method in the performance evaluation of the IFRF and other two crisp algorithms.

The experimental results are shown in Table VIII. Similarly, the non-parametric test is performed on the results as shown in Table IX.

TABLE VIII
MEAN ACCURACY RESULTS OBTAINED BY IFRF, ADABOOST AND RF IN NOISE DATASETS

| Datasets | No Noise | | | Moderate Noise (10%) | | | Heavy Noise (30%) | | |
|---|---|---|---|---|---|---|---|---|---|
| | IFRF | AdaBoost | RF | IFRF | AdaBoost | RF | IFRF | AdaBoost | RF |
| APP | 90.56 | 90.68 | 92.12 | 90.55 | 88.95 | 90.49 | 76.37 | 76.31 | 80.22 |
| AUS | 86.67 | 87.39 | 87.54 | 85.36 | 82.32 | 86.96 | 85.22 | 65.94 | 81.30 |
| BAN | 73.48 | 81.44 | 77.74 | 68.27 | 75.34 | 72.54 | 66.22 | 63.65 | 68.47 |
| DER | 98.09 | 97.81 | 98.63 | 81.16 | 94.53 | 97.54 | 71.87 | 74.86 | 92.91 |
| GLA | 78.47 | 79.90 | 81.31 | 75.68 | 74.76 | 72.40 | 64.97 | 59.38 | 64.47 |
| HAY | 84.37 | 80.63 | 85.63 | 75.63 | 66.25 | 81.25 | 64.38 | 59.38 | 67.50 |
| IRI | 97.33 | 94.67 | 96.66 | 94.00 | 84.67 | 94.00 | 94.00 | 66.00 | 82.67 |
| MAG | 86.03 | 87.27 | 86.20 | 83.55 | 85.36 | 83.70 | 83.56 | 82.71 | 85.08 |
| MAM | 84.07 | 80.64 | 83.45 | 82.83 | 75.76 | 82.21 | 81.68 | 68.37 | 80.64 |
| NEW | 98.14 | 93.02 | 96.28 | 93.95 | 93.02 | 94.88 | 81.40 | 77.21 | 82.79 |
| RIN | 95.89 | 96.91 | 93.12 | 90.93 | 92.96 | 90.77 | 90.59 | 78.64 | 90.15 |
| SEG | 97.48 | 98.66 | 96.71 | 94.97 | 94.24 | 93.03 | 92.81 | 74.72 | 92.55 |
| TAE | 57.59 | 64.34 | 60.92 | 53.61 | 64.28 | 57.59 | 50.28 | 54.32 | 52.99 |
| VEH | 74.23 | 78.73 | 75.53 | 71.16 | 73.41 | 72.22 | 67.62 | 61.71 | 70.45 |
| VOW | 97.98 | 97.17 | 92.73 | 89.09 | 88.99 | 76.67 | 72.32 | 70.10 | 71.62 |
| WDC | 97.54 | 95.78 | 96.13 | 95.60 | 95.96 | 94.90 | 94.38 | 85.23 | 90.34 |
| WIN | 99.44 | 93.83 | 98.29 | 98.30 | 95.48 | 96.63 | 91.00 | 79.76 | 86.51 |
| WIS | 97.42 | 96.71 | 96.99 | 96.42 | 91.85 | 96.28 | 95.28 | 78.54 | 92.42 |

TABLE IX
RESULTS OF THE FRIEDMAN TEST ON THE ACCURACY OBTAINED BY IFRF, ADABOOST AND RF IN NOISE DATASETS

| Algorithms | No Noise | | Moderate Noise (10%) | | Heavy Noise (30%) | |
|---|---|---|---|---|---|---|
| | Mean Rank | p-value | Mean Rank | p-value | Mean Rank | p-value |
| IFRF | 2.00 | 0.801 | 1.86 | 0.744 | 1.56 | <0.001 |
| RF | 1.89 | | 2.03 | | 1.61 | |
| AdaBoost | 2.11 | | 2.11 | | 2.83 | |

TABLE X
RESULTS OF THE PAIRWISE FRIEDMAN TEST ON IFRE AND OTHER TWO CRISP ALGORITHMS WITH HOLM METHOD IN HEAVY NOISE DATASETS
(SIGNIFICANCE LEVEL = 0.05)

| Algorithms | Holm | p-value | Hypothesis |
|---|---|---|---|
| RF | 0.025 | **0.637** | Not rejected |
| AdaBoost | 0.05 | **0.001** | Reject |

It can be seen from the experimental results that IFRF and RF perform comparable on the moderate noise dataset. AdaBoost performance is not good. On the heavy noise dataset, IFRF performs slightly better than RF, and the hypothesis test results also prove that there are significant differences between the methods. Therefore, we believe that under moderate noise, RF and IFRF have similar performance; under heavy noise, IFRF performs slightly better than RF. And IFRF performs obviously better than AdaBoost, which shown in Table X.

## V. CONCLUSION

In this study, we propose a new random forest ensemble based on intuitionistic fuzzy decision trees. We extend the existing fuzzy decision trees to the intuitionistic fuzzy theory and advance to more effective modeling the classification trees based on fuzzy sets. The intuitionistic fuzzy sets are more sophisticated than the traditional fuzzy description of data by incorporating the hesitation as the degree of uncertainty. In the proposed IFDT, the hesitation parameter is considered in the non-membership function and in the information transmission process of decision trees. It takes into account the relationship between the value and the concept, which is reflected by the degree of membership and non-membership. The proposed IFDT can also deal with imperfect datasets that contain missing and fuzzy values.

Based on the proposed IFDT, we further develop the IFRF with two schemes of combination and an ensemble method. It is well established in the literature that ensemble methods can increase the classification accuracy compared to individual classifiers. In the proposed IFRF, features are randomly selected when splitting nodes using bootstrapped samples. Such randomness contributes to the robustness property of the random forest ensemble and improves the diversity of the base trees. IFRF calculates the intuitionistic fuzzy entropy of each sample, and uses the intuitionistic fuzzy information gain method for feature selection in nodes. In the information transferring process between nodes, the transfer of non-membership degrees is also added, increasing the differences between different trees in a forest. In addition, IFRF supports multi-split of nodes, which improves the fit of the dataset. In general, it has the ability to increase variance and reduce bias, enhancing the performance of bagging ensemble classifier.

For computation, we develop algorithms for fitting the proposed IFDT and IFRF and implement them using Python. We also conduct numerical studies using public datasets and compare the proposed IFDT and IFRF to multiple alternative methods. Competitive and superior performance in accuracy is observed. Furthermore, the proposed algorithm can also be applied to specific data containing hesitation information, such as voting classification, human decision, etc.

Our study also has limitations. First, the proposed method is based on the trapezoid intuitionistic fuzzy membership function and can be extended to more complex membership functions. Second, we adopt the K-means method to discretize the domains of the fuzzy sets due to its simplicity. There exist other methods for fuzzy clustering, such as fuzzy C-means. Given the scope of this study, we choose K-means for fuzzy partition and achieve satisfactory results with moderate computational cost. Finally, the two voting schemes for the random forest ensemble are not explicitly compared. More numerical studies are needed for future work.

## APPENDIX A
### TUNNING OF HYPERPARAMETERS

In order to study the influence of each parameter on the performance of the IFDT algorithm, we select three commonly used datasets and adjust each hyperparameter separately to assess the effect of the hyperparameter on the algorithm performance, which provides a basis for the optimization of the hyperparameter. We denote the number of clusters as $C$, the shape of the membership function as $S$, the hesitation parameter as $d_\pi$, the maximum depth as $d_\beta$, the minimum split sample number as $n_\alpha$, and the number of trees as $n_\gamma$. The default values of each hyperparameter at initialization is shown in Table XI.

TABLE A XI
HYPERPARAMETER DEFAULT VALUES

| Hyperparameter | $C$ | $d_\pi$ | $S$ | $d_\beta$ | $n_\alpha$ | $n_\gamma$ |
|---|---|---|---|---|---|---|
| Default value | 2 | 0.9 | 7 | 5 | 5 | 100 |

The first step is to tune $C$. This parameter determines the number of child nodes that split at each node. As shown in Figure A5, it can be found that $C$ has a great impact on the performance of the model. On the glass dataset, a 40% difference between the best-performing $C$-value and the worst $C$-value result is observed. The best-performing $C$-value is also different on different datasets. Therefore, the default value is set to the minimum value of 2. When fitting each data set, it is necessary to tune the $C$-value to select the best model.

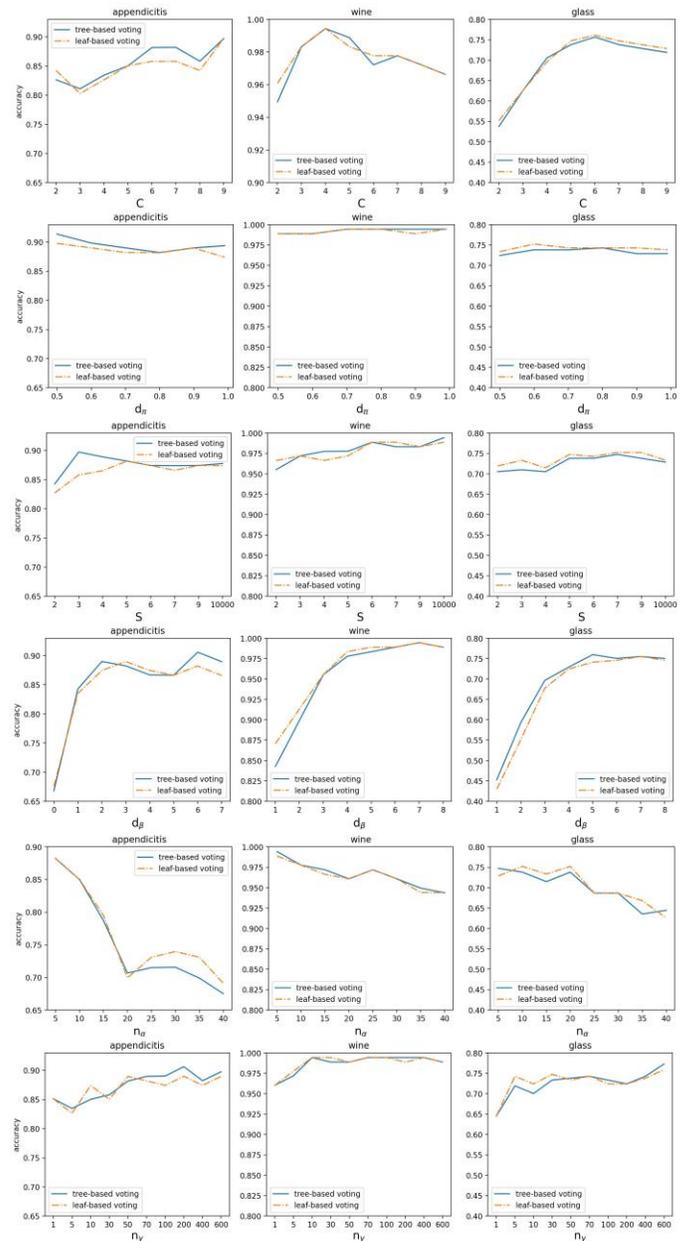

Fig. A5. Results of the tunning for the IFRF

In the following experiments, we set $C$ to the best value found in the first experiment for each dataset, while the other hyperparameters are left at their default values. We then tune them one by one.

When tuning $S$, the magnitude of the change in the prediction results is much smaller than that of $C$. In the fitting process of the glass data set, $S$ has shown satisfactory performance over a large interval. Therefore, compared to $C$, $S$ is less sensitive in model performance, and a larger step size can be used in searching its best value. For instance, the step size can be set to be 0.2. The default value is set to be 0.9.

The prediction results change little as we change the value of $d_\pi$. Increasing $d_\pi$, the performance in the three datasets becomes slightly better. We set the default value to be 0.9. It can be considered that this hyperparameter has the least impact on the model, and it is generally set to its default value.

In tuning $d_\beta$, in all datasets, as $d_\beta$ increases, the classification results gradually become better, and finally reach a stable state without obvious overfitting. When $d_\beta=5$, all model results converge, so the default maximum depth is 5.

In tunning $n_\gamma$, in all data sets, the performance shows a tendency to increase first and then converge. When $n_\gamma=100$, the results of the model tend to be stable on all data sets, so we set the default value to be 100.

When tunning $n_\alpha$, as the pruning hyperparameter increases, the performance of the model gradually decreases. Pruning is used to alleviate the phenomenon of overfitting, and there is no overfitting in the datasets of our experiments. With the increase in $n_\alpha$, it leads to under-fitting. We set the default value to be 5, and need to increase this parameter appropriately when overfitting.

The influence of the hyperparameters $d_\beta$, $n_\gamma$ and $n_\alpha$ relating to trees on the model results is consistent with their influence in the random forest, and the corresponding hyperparameter tuning process is similar.

When selecting hyperparameters for a large dataset, a single IFDT can be used for tuning first to select the optimal discretization hyperparameters. Afterwards, other hyper-parameters can be selected by integrating multiple random IFDT learners, which can greatly reduce the computational cost.

## APPENDIX B
## COMPLEXITY OF IFRF

Assuming that the dataset has n samples, m features, and all features are numerical data, each ensemble algorithm contains t base learners, and no pruning operation is performed.

For a random forest, training a tree first requires sorting each feature. Sorting one feature requires the time complexity to be O(n*log n), so sorting all features requires O(m*n*log n). During the crisp decision tree training process, we need to calculate information gain at each threshold. For one feature, it takes O(n) and O(n*m) for m features. A total of *t* base learners are trained and the total cost is O(t*d*n), so the total time complexity for random forest is O(m*n*log n) + O(t*m*n).

Since m is generally much smaller than n in the real dataset, big O notation is adopted for the time complexity O(m*n*log n) + O(t*m*n).

For an intuitionistic fuzzy random forest, the data needs to be intuitionally fuzzificated first. The center point is obtained using K-means, and the time complexity is O(m*n). After that, the intuitionistic fuzzy assignment and the calculation of intuitionistic fuzzy entropy are performed on each sample. The cost of each sample is O(m) and the total cost is O(m*n).

In the intuitionistic fuzzy decision tree training process, we don't need to calculate each threshold. We only need to calculate the intuitionistic fuzzy information gain of each feature to complete the split, which takes O(m). Training *t* base learners costs O(t*m). Thus, the total time complexity for the intuitionistic fuzzy random forest is O(m*n) + O(t*m).

Intuitionistic fuzzy random forest training time complexity is less than that of random forest.

In terms of space complexity, two algorithms are both O(p), where p is the number of nodes in all trees. Since the intuitionistic fuzzy decision tree supports the splitting of multi-nodes, the number of nodes trained by the intuitionistic fuzzy tree may be greater than that of the crisp decision tree based on binary tree splitting. However, since only one split feature name is stored in each node, we believe that this space complexity is acceptable.

In the actual test, a txt file is used to store 100 intuitionistic fuzzy decision trees with a maximum depth of 8. When the number of sub-nodes of each node split is 2, 4, 6, 8, and 10, the required storage space is 0.38MB. 4.51MB, 5.33MB, 4.03MB and 2.81MB. Even if the number of intuitionistic fuzzy decision trees fitted in the algorithm is further increased, the memory usage would not exceed 30MB. Given that the current mainstream computer memory is greater than 8GB, we think this memory usage is completely acceptable.

With the increase in the number of split nodes, the required storage space shows a trend of rising first and then falling. The peak of space occupied probably reaches its maximum when the number of split nodes is 6. As the number of split nodes increases, more samples will be divided into multiple child nodes, making it more difficult for child nodes to trigger pruning conditions. However, as the split node increases to a certain threshold, the phenomenon will also reach a peak. As the number of split nodes increases, the average number of samples assigned to each child node gradually decreases, making it easier for child nodes to trigger the pruning condition, resulting in a reduction in the number of final nodes. More customized pruning methods for the intuitionistic fuzzy random forest can be designed in the future.